\documentclass[a4paper,11pt]{article}
\usepackage[a4paper, total={6in, 8in}]{geometry}

\usepackage{url}
\usepackage[hidelinks]{hyperref}
\usepackage{xcolor}
\usepackage{amsmath,amssymb,tikz-cd}
\graphicspath{ {./images/} }

\usepackage{eucal, bm}
\newcommand{\catBold}[1]{{\normalfont\textbf{{#1}}}}
\newcommand{\cat}{\mathcal}
\newcommand{\ob}[1]{\mathrm{Ob}(#1)}

\title{\LARGE \bf
Category Theory for Autonomous Robots: The Marathon 2 Use Case 
}

\author{Esther Aguado, Virgilio Gómez, Miguel Hernando, Claudio Rossi and Ricardo Sanz}
\date{%
   \footnotesize{ Universidad Politécnica de Madrid, Autonomous Systems Laboratory \\ and Center for Automation and Robotics; c/ José Gutierrez Abascal 2, 28006 Madrid, Spain\\    {\texttt{\{e.aguado, virgilio.gomez.lambo, miguel.hernando, claudio.rossi, ricardo.sanz\}@upm.es}}}}

\begin{document}

\maketitle

\begin{abstract}
Model-based systems engineering (MBSE) is a methodology that exploits system representation during the entire system life-cycle. The use of formal models has gained momentum in robotics engineering over the past few years. Models play a crucial role in robot design; they serve as the basis for achieving holistic properties, such as functional reliability or adaptive resilience, and facilitate the automated production of modules. We propose the use of formal conceptualizations beyond the engineering phase, providing accurate models that can be leveraged at runtime. This paper explores the use of Category Theory, a mathematical framework for describing abstractions, as a formal language to produce such robot models. To showcase its practical application, we present a concrete example based on the Marathon 2 experiment. Here, we illustrate the potential of formalizing systems\textemdash including their recovery mechanisms\textemdash which allows engineers to design more trustworthy autonomous robots. This, in turn, enhances their dependability and performance.
\end{abstract}

\section{Introduction}

Autonomous robot engineering constitutes a major challenge in system development\textemdash design, construction, assurance, operation\textemdash due to their complexity in both hardware/software and operational environments. Their construction requires well-orchestrated and knowledge-intensive engineering processes that can benefit from rigorous \emph{Systems Engineering} (SE) practices. Having a suitable technology to develop \emph{system models} is a major step in solving this challenge.


We use models to abstract and manage system complexity. Model-based systems engineering (MBSE) is a semi-formalized methodology that exploits a system representation with reduced detail so that its structure and behavior can be managed. MBSE provides support for requirement definition, design, analysis, verification, and validation throughout the entire life-cycle of intricate systems.

Of all the aspects of the system, a highly critical one is \emph{system architecture}~\cite{Bass-2021}, because it affects and even determines most of the properties of the system. MBSE modeling of system architectures is a cornerstone of reliable system construction. There are languages that support MBSE, such as AADL~\cite{aadl} or SysML~\cite{sysml}, which facilitate the definition, communication, and maintenance of system models. These languages are typically used to document and reason about the properties of the system and its consistency. However, when systems are deployed in unpredictable environments, it is hard to ensure its behavior at the development time.



We propose the use of \emph{Category Theory} (CT) as a framework to support system modeling and behavioral analysis for complex Systems of Systems (SoS). CT is a general theory of mathematical structures. It was invented in the 1940s to unify and synthesize different areas in mathematics~\cite{spivak-2014}. It can be seen as a tool set for describing general structures and maintaining control over which aspects are preserved when performing abstractions. Moreover, its mathematical foundation enables powerful communication between apparently unrelated fields.

Our motivation is based on the difficulties in addressing the reality gap between robot usage in a controlled environment and its deployment in open, real-world scenarios. We aim to address dependability using robust models at runtime as a tool to better understand the situation and react properly in unstructured environments. In this article, we draw a landscape of how CT can be used for this purpose. In particular, the main contributions are as follows. 

\begin{itemize}
    \item A brief guide on how to use CT to design systems,
    \item its application to a state-of-the-art mobile robot, and
    \item a discussion on the benefits of using CT in robotics.
\end{itemize}

Beyond this introduction, the paper is organized as follows. Section \ref{sec:concepts} addresses the description of some fundamental CT elements that can add value in robotic engineering processes. Section \ref{sec:relatedwork} describes some related work on applied CT in domains of close interest for roboticists\textemdash databases, knowledge management, hierarchical systems, robotics, etc. Section \ref{sec:usecase} describes a concrete use case: the Marathon 2 experiment on autonomous mobile robots in real social environments. Section \ref{sec:model} describes the use of CT in the Marathon 2 use case. Finally, Section \ref{sec:discussion} discusses the potential use of CT in the modeling of robotic systems in an MBSE context. The paper ends with some conclusions, acknowledgments, and references.

\section{Category Theory Concepts for Robotics}\label{sec:concepts}

In this Section we define the specific language we will use to our CT-based system model. CT allows us to be completely precise about otherwise informal concepts~\cite{Kozen2006}. We start with some basic elements to represent structures and evolve them into more general formalisms, \emph{operads}. Lastly, we introduce the concept of \emph{pushout}, which provides a method to condense and combine data from pre-existing structures.

Note that this introduction is just a basic description for robotics; for further reading, refer to~\cite{lawvere, lane, awodey} from a mathematical perspective,~\cite{barr} from a computer science perspective, and~\cite{spivak-2014, fong2019} for a scientific application perspective with a strong mathematical flavor. 

\subsection{Basis: Category, Functor, and Natural Transformation}\label{sec:basis}

A \emph{category} $\cat{C}$, is a collection of elements with relations between them. It constitutes an aggregation of objects with an imposed structure~\cite{Lloyd2021}. To specify a category, we need three constituents:
\begin{itemize}
    \item A collection of \emph{objects}, $\ob{\cat{C}}$,
    \item a collection of \emph{morphisms} $f: X \to Y$, for every pair of objects $X, Y \in \ob{\cat{C}}$, and
    \item a \emph{composition} operation $g \circ f: X \to Z$ composing morphisms $f: X \to Y$ and $g: Y \to Z$.  
\end{itemize}

Additionally, categories must satisfy the following conditions:
\begin{itemize}
    \item identity: for every object $X \in \ob{\cat{C}}$, exists an identity morphism $id_X: X \to X$.
    \item associativity: for any three morphisms, $f: X \to Y$, $g: Y \to Z$, $h: Z \to W$, the following expressions are equal:  $(h \circ g) \circ f =  h \circ (g \circ f) =  h \circ g \circ f$
    \item unitality: for any morphism $f: X \to Y$, the composition with the identity morphisms at each object does not affect the result, $id_X \circ f = f$ and $f \circ id_Y = f$.
\end{itemize}

A simple example is the $\catBold{Set}$ category, in which objects are sets, morphisms are functions between sets, and a composition operation is the composition between its functions.

A \emph{functor} $F$, is a map between two categories $\cat{C}$, $\cat{D}$. It assigns objects to objects and morphisms to morphisms, preserving identities and composition properties. Functors preserve structures when projecting one category inside another. 



Functors can map between the same types of category, such as $\catBold{Set} \to \catBold{Set}$, or between different categories, such as $\catBold{Set} \to \catBold{Vect}$, between the category of sets and the category of vector spaces.

A \emph{natural transformation} $\alpha$, is a structure-preserving mapping between functors. Functors project images of a category inside another; whereas natural transformations shift the projection defined by a functor $F$ into the projection defined by a functor $G$. Diagram \ref{eq:natural_transformation} relates these three concepts. There are two categories $\cat{C}, \cat{D}$ and two different functors $F, G: \cat{C} \to \cat{D}$. The two functors are linked by the natural transformation $\alpha: F \Rightarrow G$.

\begin{equation}\label{eq:natural_transformation}
\begin{tikzcd}[column sep=huge, row sep=huge]
\mathcal{C}
  \arrow[bend left=50]{r}[name=C,label=above:$\scriptstyle\mathrm{F}$]{}
  \arrow[bend right=50]{r}[name=D,label=below:$\scriptstyle G$]{} &
\mathcal{D}
  \arrow[shorten <=10pt,shorten >=10pt,Rightarrow,to path={(C) -- node[label=right:$\alpha$] {} (D)}]{}
\end{tikzcd}
\end{equation}

To specify a natural transformation, we define a morphism $\alpha_c: F(c) \to G(c)$ for each object $c \in \cat{C}$, such that for every morphism $f: c \to d$ in $\cat{C}$ the composition rule $\alpha_d \circ F(f) = G(f) \circ \alpha_c$ holds. This condition is often expressed as the commutative diagram shown in Diagram \ref{eq:natural_transf_commutative}, where the natural transformation morphisms are represented as dashed arrows. This means that the projection of $\cat{C}$ in $\cat{D}$ through $F$ can be transformed into projections through $G$. The commutative condition implies that the order in which we apply the transformation does not matter. 

\begin{equation}\label{eq:natural_transf_commutative}
\begin{tikzcd}[row sep=huge]
F(c) \arrow[d, "F(f)"'] \arrow[r, "\alpha_c", dashed] & G(c) \arrow[d, "G(f)"] \\
F(d) \arrow[r, "\alpha_d", dashed]                   & G(d)                  
\end{tikzcd}
\end{equation}
\subsection{Representing Resources: Wiring Diagrams}\label{sec:wiring}

Fong and Spivak ~\cite{fong2019} define \textit{wiring diagrams} as an answer to the question ``Can I transform what I have into what I want?'' Wiring diagrams constitute a CT-formalization of engineering block diagrams. In these diagrams, boxes represent transformations (i.e. morphisms in a category), wires represent resources (i.e. objects in a category), and two boxes in series are the result of composing morphisms.

In the broad sense, wiring diagrams depict monoidal categories $(\cat{C}, \otimes, \{1\})$, which are categories $\cat{C}$ equipped with a tensor product $\otimes$ and a monoidal unit $\{1\}$ that allows the combination of objects in the category. This structure provides a rationale to place boxes in parallel in the diagram. Lastly, \textit{symmetric monoidal categories} (SMC) and its respective coherence conditions support the use of feedback wires and swapping in these diagrams. A precise definition of wiring diagrams can be found in~\cite{lane} and~\cite{Selinger2010}.

\subsection{Representing Networks: Operads and Algebras}\label{sec:operads}
Categories, functors, and natural transformations are the building blocks of CT. They allow us to express objects, relationships among them, and abstractions while preserving their structure. This approach imposes some directionality since morphisms, functors, and natural transformations are maps with a domain and a co-domain. From an engineering perspective, we could see domains and co-domains as input and output ports, respectively. However, we sometimes prefer to express systems in terms of networks in which ports exit without a direction. CT provides a generalization over categories, operads, to model assemblies of structures.

An \emph{operad} is a mathematical object that describes operations with multiple inputs and one output~\cite{Yau2018}; classically, it only describes one object. We use here what are commonly known as colored operads, which can hold many objects (i.e. colors). Leinster~\cite{leinster2004} provides a precise definition of operads. An operad consists of:
\begin{itemize}
    \item a set of objects $\ob{\cat{O}} = X_1, \ldots, X_n$ each of which can be seen as cells,
    \item a set of morphisms $\varphi, \psi, \ldots$ that constitute the operations and act as an assembly of cell interfaces creating an external interface, and
    \item a composition formula $\circ$ or substitution, that enables the assembly of morphisms, i.e. the composition of interfaces.
\end{itemize}



Operads allow us to define abstract operations, but we need to ``fill'' those cells to ground them. An \emph{algebra} is an operad functor from the operad to the $\catBold{Set}$ category, $F: \cat{O} \to \catBold{Set}$ that produces a concrete realization. It is subject to the operations and composition relations specified by the operad. In practice, an algebra determines:

\begin{itemize}
    \item for any object, $x \in \cat{O}$, a set $F(x)$,
    \item for any morphism in the operad, $f: (x_1, \ldots, x_n) \to y$ and for any element $f_i \in F(x)$, a new object $f' = F(f) (f_1, \ldots, f_n) \in F(y)$. 
\end{itemize}

Different algebras allow us to ``fill'' the operad structure from different perspectives and then combine them using a morphism between algebras $\delta: A \to B$. In Section \ref{sec:model}, \emph{we propose two algebras, structure and behavior, for the system operad}. These algebras can be seen as the semantics of the syntax established in the operad.

\subsection{Combining information: Pushouts}
So far we have introduced the machinery to focus on roles and structure via categories and operads. In this work, our aim is to use abstractions to build the best possible system for each situation. Pushouts serve as a valuable tool in this process as they provide the best approximation of an object in a category that satisfies certain conditions.

\emph{Pushouts} can be seen as a way to combine two objects in a category with a common object so that all information is preserved. For example, in a category $\cat{C}$ with three sets as objects, $X, Y, A \in \ob{\cat{C}}$, we can have a set $A$ representing three robots, $A = \{robot\_1, robot\_2, robot\_3\}$, a set $X$ representing robot grippers, $X = \{finger\_gripper,\\ vacuum\_gripper\}$, and a set $Y$ representing the maximum speed (m/s) in wheeled robots, $Y = \{1.5, 3\}$. There are two morphisms in our category, $f: A \to X$ assigning $robot\_1$ a maximum speed of $1.5$ m/s and $robot\_2$ a maximum speed of $3$ m/s; and morphism $g: A \to Y$ assigning  $robot\_2$ to the $three\_fingers\_gripper$ and $robot\_3$ to the $vacuum\_gripper$.

The pushout $X_{+A}Y$ can be seen as the summary of objects and its relationships, in a $\catBold{Set}$ category, a non-disjoint union. In this case, a set $X_{+A}Y = \{ robot\_1\_speed\_1.5, robot\_2\_$\\$speed\_3\_and\_finger\_gripper, robot\_3\_vacuum\_gripper\}$ and its corresponding morphisms to sets $X, Y$. Note that objects connected by morphisms are seen as the same element, i.e. the same system made up of subsystems $X, Y, A$.

Pushouts are usually expressed as Diagram \ref{eq:pushout1} in which $f', g'$ are the pushout morphisms that respectively link $Y, X$ with the pushout object $X_{+A}Y$. If this diagram commutes, i.e. $g' \circ f = f' \circ g$, there exists a unique morphism $u: X_{+A}Y \to T$ such that Diagram \ref{eq:pushout2} also commutes. The dashed arrow represents the unique morphism, and the element $T$ represents the unique pushout object. $l_x, l_y$ constitute the validity limits of $T$~\cite{Lloyd2010}.

\begin{equation}\label{eq:pushout1}
\begin{tikzcd}[row sep=large]
A \arrow[r, "g"] \arrow[d, "f"'] & Y \arrow[d, "f'"] \\
X \arrow[r, "g'"']               & X_{+A}Y                   
\end{tikzcd}
\end{equation}

\begin{equation}\label{eq:pushout2}
\begin{tikzcd} [row sep=large]
A \arrow[r, "g"] \arrow[d, "f"']                   & Y \arrow[d, "f'"] \arrow[rdd, "l_y", bend left] &   \\
X \arrow[r, "g'"'] \arrow[rrd, "l_x"', bend right] & X_{+A}Y \arrow[rd, "u", dashed]                 &   \\
                                                   &                                                 & T
\end{tikzcd}
\end{equation}

\section{Related Work}\label{sec:relatedwork}
CT has proven to be a powerful tool for abstract reasoning about mathematical structures. However, it is still not as widely used in applied scenarios as other mathematical disciplines, such as linear algebra or calculus. The main reasons could be that it is a too specialized and abstract field, with a steep learning curve, and the community. Besides its main uses in mathematics, it is applied in computer science and software engineering; mainly to identify commonalities and patterns across different domains. A notable contribution is its use in the design and implementation of the C++ Standard Template Library~\cite{stepanov2014}. 

In recent years, the number of fields that implement CT-based approaches has increased. In knowledge representation, Spivak~\cite{spivak2012} proposes the ontology log, \emph{olog}, a CT-formalism similar to relational database schemas. It provides a friendly definition language and supports safe data migration~\cite{spivak2015}. Aligned with this approach, Patterson~\cite{Patterson2017} introduces a relational ontology, which seeks to bring ologs closer to description logics. These works rely on categories, functors, and natural transformations and use pushouts to combine information and generate new concepts from old ones.

From the robotics perspective, Censi~\cite{Censi16} introduces the theory of co-design to optimize multi-objective systems. Zardini~\cite{Zardini2021} applies interconnected co-design problems to a self-driving car. This approach applies wiring diagrams to optimize resource allocation in the design phase. 

Perhaps the vision of Lloyd is more in line with our work. Lloyd~\cite{Lloyd2021} suggests the use of CT language and theory to study systems and reflexes on its implications. Its domain application is biological sciences. In~\cite{Lloyd2010}, he provides a CT-centered approach to model and simulate the emergent properties of intercellular interaction.

Like us, Bakirtzis~\cite{Bakirtzis2021} studies the unification of requirements, behaviors, and architectures for complex systems. He proposes the use of wiring diagrams as an alternative to engineering modeling languages such as SysML and provides a grounding example on an unmanned aerial vehicle. Its perspective on \emph{contracted behaviors}, which composes behaviors and requirements through an algebra, has inspired our recovery model introduced in Section \ref{sec:CTmodelrecov}. The main limitation of this approach is the directionality imposed by wiring diagrams. 

The generalization provided by operads appears to be a better structure to model complex systems. Schweiker et al.~\cite{Schweiker2015} propose the application of wiring diagrams operad to model safe critical systems. In particular, they establish viewpoints with a sound mathematical framework in which behaviors and error propagation can be assessed from different perspectives. This model is applied to evaluate the failure effects of air traffic management communication channels on aircraft control. Our approach is based on the same abstract construction, the operad of wiring diagrams, as fundamental structure for the system model. 

Breiner et al.~\cite{Breiner2020} offer another perspective on using operads to model systems. It uses the separation of concerns provided by operads to analyze a length calibration device, the length scale interferometer (LSI), used at the US National Institute of Standards and Technology (NIST). They model the system interfaces as a port graph operad. Then, they use a functor between the system and the probability of failure to perform the failure diagnosis. Foley et al.~\cite{foley2021} examine operad-based problem modeling from two angles. It takes a top-down modeling approach using wiring diagrams operad to analyze the LSI case from~\cite{Breiner2020}, and a down-top perspective using network operad to design a search and rescue architecture and its design mission task plan.

These works show that CT is seen as a tool to increase the dependability and explainability of the system, which is a major concern in any technology-centered domain. The growing number of publications suggests the potential of CT for practical use. However, there are still challenges in bridging the abstraction provided by CT and the concrete implementation in real-world scenarios. In this work, we try to take a step in that direction, defining a model that can be used both at the design phase and during robot operation.

\section{The Marathon 2 Use Case}\label{sec:usecase}

The Marathon 2 experiment~\cite{macenski2020} proposes a guidance, navigation, and control (GNC) solution for mobile robots, which has become a widely adopted stack for handling perception, mapping, localization, path planning, and control. This software, known as \emph{Nav2}, provides a design pattern that seeks modularity, flexibility, and extensibility. Section \ref{sec:model} models the system deployed in this experiment using CT.

The described experiment intends to cover a distance greater than a marathon without human assistance. Specifically, a robot shall traverse 16 waypoints several times in a university setup. In operation, the robot faced difficulties such as human crowds and navigation around complex areas such as a central staircase, a high-traffic bridge, a hallway, and a narrow doorway.

The robots used in the experiments, a Tiago~\cite{pal2022} and a RB-1~\cite{robotnik_2022} have similar base dimensions and their speed was limited to 0.45 m/s for safety reasons, which is below their maximum speed. They use differential steering on a circular base, as required by the implementation of the Adaptive Monte Carlo Localization (AMCL)~\cite{fox2001} and the A$^*$ planner~\cite{Konolige2000}. These robots use a laser and a depth camera as the main sensors. 

\begin{figure}[ht]
    \centering
    \includegraphics[width=0.75\textwidth]{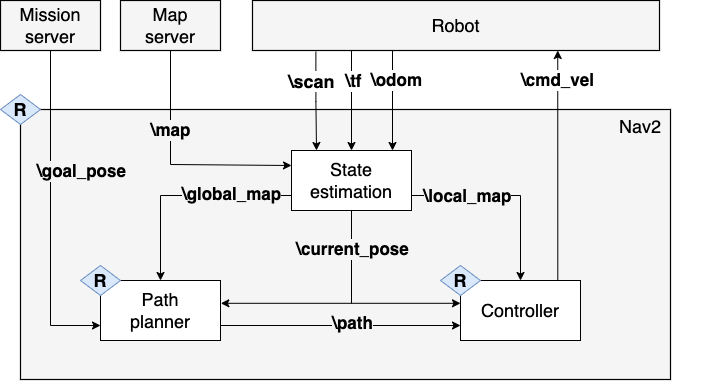}
    \caption{Marathon 2 software architecture. ROS topics are depicted in bold and starts with a ``$\setminus$'' symbol. The gray boxes are the main elements: mission server, map server, robot and Nav2. The Nav2 subsystem is further decomposed into the white boxes: state estimator or navigator, path planner or guidance, and controller. Note that the path planner, the controller and the Nav2 boxes have a diamond shape in the corner with an ``R'' representing that exists available recovery actions for such subsystems.}
    \label{fig:marathon2}
\end{figure}

Figure \ref{fig:marathon2} depicts the Marathon 2 architecture. It enables a mobile robot to autonomously reach a goal state, such as a specific position and orientation relative to a specific map. It takes as input a map, a goal pose, and the current pose, and provides velocity commands to drive the robot. In the experiments, the goal pose was the sequence of 16 waypoints. The software relies on a behavior tree (BT)~\cite{Colledanchise_2018} to provide flexibility and adaptability to perform the task. In this case, it coordinates (i) guidance through a path planner to find the best route to the goal pose, (ii) navigation based on AMCL for state estimation, and (iii) a controller to compute the path and local information to generate a precise velocity control signal. Additionally, it provides a recovery server to solve runtime contingencies.

\subsection{Recoveries in Marathon 2}\label{sec:recoveries}

Marathon 2 uses ad-hoc solutions to handle contingencies. Recovery actions depend on the specific failing component and the data it manages. As shown in Figure \ref{fig:marathon2}, Marathon 2 provides recoveries for the path planner, the controller, and the entire Nav2 subsystem. Nav2 logic uses a BT to orchestrate the execution of tasks. Recoveries are encoded as BT fallbacks, triggered when the specific task fails. In particular, (i) when the system cannot generate a plan, it clears the global map used by the path planner; (ii) when the controller fails to determine a suitable velocity, it clears the corresponding area in the environmental model (i.e. local map); and (iii) when the whole GNC process fails, it first clears the entire environmental model, then spins the robot in place to reorient local obstacles, and ultimately waits a timeout.

Macenski et al.~\cite{macenski2020} identified two primary scenarios that triggered recovery behaviors in the robot. The first scenario involved crowded spaces where the robot was unable to compute a clear path to its destination. If the \emph{clear map recovery} approach failed and the path was obstructed by people, the robot used \emph{wait recovery}, which paused navigation until the path was clear. The second scenario occurred when localization confidence was low, particularly in long corridors with repetitive features and many people. In these cases, the \emph{spin recovery} approach\textemdash in which the robot spins in place\textemdash was used to improve the confidence in position before resuming its task.

While these ad-hoc recovery approaches did utilize the knowledge and experience of engineers in dealing with contingencies, they lacked the ability to provide a clear rationale for the robot's actions. As a result, the reasoning behind these approaches remained solely in the minds of the engineers. This lack of transparency can give rise to concerns about the reliability of the system, as a particular recovery approach may not be appropriate for a given situation or may even adversely affect a properly functioning subsystem. For example, the planner recovery approach, which clears the global map, can propagate to the local map utilized by the faultless controller.
\section{CT-model for the Marathon 2}\label{sec:model}

In this Section, we formalize the Marathon 2 use case in a model based on CT with three intentions: (i) provide a formal model of the system with a strong mathematical foundation, (ii) represent all valid equivalent designs, and (iii) exploit that model at runtime, i.e. implement sound model-based recoveries.

\subsection{System model}

Operads offer an effective modeling procedure for hierarchical structures. We base our model on the wiring diagram operad, $\cat{W}$, similar to the wiring diagram operad for dynamical systems explicitly defined in~\cite{Vagner2014}. Our building blocks, i.e. operad objects, are boxes $x \in \ob{\cat{W}}$ with two types of interface: resources required and resources provided. The operad morphisms $\varphi, \psi, \ldots$ represent which composite interfaces can be produced out of atomic ones, as explained in Section \ref{sec:wiring}. The composition formula $\circ$ ensures that morphisms can be composed creating higher-level structures in terms of resources.

In Section \ref{sec:operads} we claim that operads define abstract operations, they act as ``placeholders'' to create a theory of composition~\cite{spivak-2014}. We use algebras to define the ``fillings'', the concrete elements that provide and consume resources. In robotics, those resources can be seen from three perspectives (i) capability, (ii) structure, and (iii) behavior.  Each viewpoint constitutes an algebra, a way to provide meaning to the abstract structure.

\subsubsection{Capability model}

Capability refers to the ``potential'' ability of a system, i.e. what it is intended to do to satisfy a need. Its algebra, $C: \cat{W} \to \catBold{Set}$, gives meaning to resources in terms of needs. In particular, the capability algebra produces an instance of a $\catBold{Set}$ category in which:
\begin{itemize}
    \item For every object $x \in \ob{\cat{W}}$, the set $C(x)$ is the set of capabilities that provides and consumes resources from $x$. 
    \item For every morphism in the operad $f: (x_1, \ldots, x_n) \to y$ and for every choice of capability $c_1 \in C(x_1), \ldots, c_n \in C(x_n)$ it creates a composed capability $c' = C(f) (c_1, \ldots c_n) \in C(y)$. 
\end{itemize}

\begin{figure}[ht]
    \centering
    \includegraphics[width=0.7\textwidth]{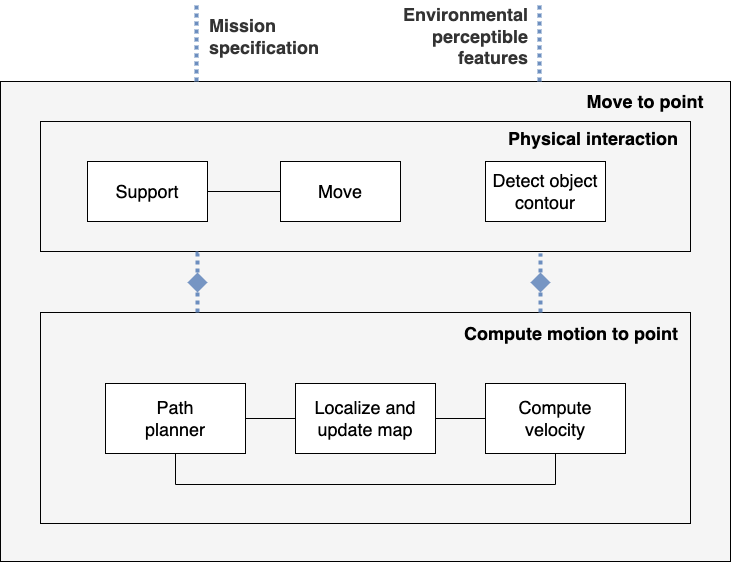}
    \caption{Capability model of the Marathon 2 system. Atomic capabilities in white boxes, composed capabilities in gray boxes. Lines in black represent links between atomic capabilities and dotted blue lines, links between composed capabilities.}
    \label{fig:capabilities}
\end{figure}

Figure \ref{fig:capabilities} displays the main capabilities of the Marathon 2 system. The operad structure is represented as boxes, the capability model is composed of two interconnected cells corresponding to the physical interaction and the computation of motion to reach the point. The mission specification and the environmental perceptible features are external capabilities consumed by the system but depicted for completeness.

\subsubsection{Structural model}

Structure represents the logical arrangement of components\textemdash hardware or software\textemdash and its interactions. It defines the system, its composing subsystems, and its connections. The structural algebra $S: \cat{O} \to \catBold{Set}$, gives meaning to the subsystems required to transform resources. In particular, the structural algebra produces an instance of a $\catBold{Set}$ category in which:
\begin{itemize}
    \item For every object $x \in \ob{\cat{W}}$, the set $S(x)$ is the set of subsystems that provide and consume resources from $x$. 
    \item For every morphism in the operad $f: (x_1, \ldots, x_n) \to y$ and for every choice of subsystem $s_1 \in S(x_1), \ldots, s_n \in S(x_n)$ it creates a system out of subsystems $s' = S(f) (s_1, \ldots s_n) \in S(y)$. 
\end{itemize}

\begin{figure}[ht]
    \centering
    \includegraphics[width=0.65\textwidth]{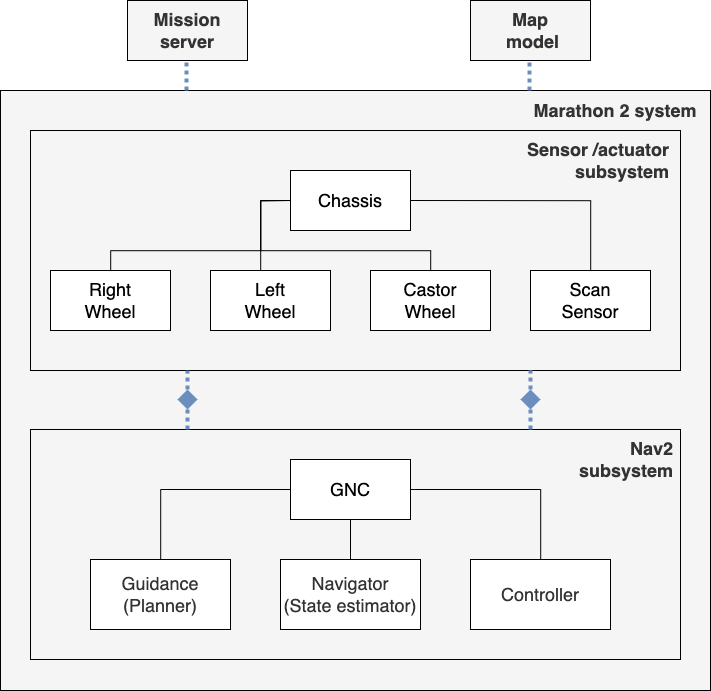}
    \caption{Structural model of the Marathon 2 system. Atomic subsystems in white boxes, composed subsystems\textemdash i.e. system of system\textemdash in gray boxes. Lines in black represent links between atomic subsystems and dotted blue lines, links between composed subsystems.}
    \label{fig:structure}
\end{figure}

In the Marathon 2 system, the structural model consists of the robot and the GNC subsystem. Its hierarchical decomposition is shown in Figure \ref{fig:structure}. The operad structure is represented as boxes, the structural model is composed of two interconnected cells corresponding to the sensor/actuator subsystem and the Nav2 subsystem. The mission server and model map are external subsystems providing resources to Marathon 2, they are depicted for completeness.

\subsubsection{Behavioral model}
Behavior is generally understood as the visible actions of a system in reaction to certain inputs, conditions, or stimuli. The behavioral algebra $B: \cat{O} \to \catBold{Set}$, gives meaning to the information flow\textemdash exchanged resources\textemdash and the resulting actions from the execution of robot commands. In particular, the behavioral algebra produces an instance of a $\catBold{Set}$ category in which:
\begin{itemize}
    \item For every object $x \in \ob{\cat{W}}$, the set $B(x)$ is the set of behaviors that provides and consumes resources\textemdash information pieces\textemdash from $x$. 
    \item For every morphism in the operad $f: (x_1, \ldots, x_n) \to y$ and for every choice of behavior $b_1 \in B(x_1), \ldots, b_n \in B(x_n)$ it creates a composed behavior, $b' = B(f) (b_1, \ldots b_n) \in B(y)$. 
\end{itemize}

A behavioral model specifies the actual response of the system in a particular situation; unlike our previous models, in which the system emerges from the combination of elements. In the behavioral model, this combination must also occur in the right order. Wiring diagrams are the most extended formalism for expressing the direction of resource consumption in SMC, as defined in Section \ref{sec:wiring}. The behavior algebra produces a finite $\catBold{Set}$ category, which constitutes an SMC~\cite{spivak-2014}, so it can be expressed as a wiring diagram similar to the Nav2 subsystem in Figure \ref{fig:marathon2}. The behavioral Marathon 2 is a wiring diagram that has (i) as boxes the concrete algorithms used according to~\cite{macenski2020}, A$^*$ planner, an AMCL state estimator, and a TEB controller, and (ii) as wires the ROS topics depicted in bold in Figure \ref{fig:marathon2}. The composite behavior takes as input a map of the environment and a goal position and drives the robot there.

\subsection{Specification of a design}

\begin{figure}[ht]
    \centering
    \includegraphics[width=0.5\textwidth]{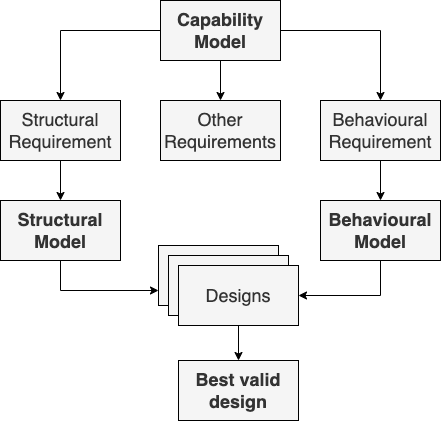}
    \caption{Relation between capability, structural and behavioral model with requirements. These models are used to establish the best valid system design.}
    \label{fig:design}
\end{figure}

A system design provides the resulting conceptualization that satisfies the three previously defined models. Figure \ref{fig:design} represents the relationship between the three concepts. Capability models refine the needs that the system must satisfy. These capabilities define requirements at different granularity. Structural and behavioral models propose solutions to meet its corresponding requirements. Solutions satisfying the three models provide the set of valid designs. Eventually, the ``best valid design'' alternative is the one to be realized.

This process can be seen as a pushout $Structure \gets Capability \to Behavior$ with requirements $SR, BR$ as morphisms between capabilities and structure or behavior. As shown in Diagram \ref{eq:design}, designs are a combination of both. If the diagram commutes, it means that it is equivalent to creating a design that first satisfies behavioral requirements and then structural ones, or vice versa. In this case, there exists a unique pushout object, known as \emph{realization}. The realization shall be the unique design that maximizes the capabilities of the system. In most systems, this is the design that maximizes behavioral performance and minimizes structural cost. For this reason, the pushout limits of validity are $max\_perf, min\_cost$.

\begin{equation}\label{eq:design}
\begin{tikzcd} [row sep=huge, column sep=large]
Capability \arrow[r, "SR"] \arrow[d, "BR"']                   & Structure \arrow[d, "BR'"] \arrow[rdd, "min\_cost", bend left] &   \\
Behavior \arrow[r, "SR'"'] \arrow[rrd, "max\_perf"', bend right] & Designs \arrow[rd, "u", dashed]                 &   \\
                                                   &                                                 & Realization
\end{tikzcd}
\end{equation}

\subsection{CT-Model-based Recoveries}\label{sec:CTmodelrecov}
Models shall be used throughout the system life-cycle. We propose the use of CT-models at runtime to reconfigure the system and increase its resiliency. Today, most subsystems have specific recovery actions to ensure proper operation at the component level. However, to fully address the open issues in dependability, the system must be aware of these atomic recoveries and how they affect other components and its shared resources. CT provides the formal framework for mapping between components and behaviors.

During robot deployment, changes in the environment or system contingencies may cause that the best designed realization is no longer valid. In this case, we propose a functor mapping to generate new \emph{valid} structural and behavioral models. These new models then evolve into a \emph{recovered} realization. 

From a CT perspective, there is a functor $\delta: S \to B$ to establish the map between structure-$\catBold{Set}$ category and behavior-$\catBold{Set}$ category resulting from its corresponding algebras. When a contingency arises, there is an alternative mapping $\delta': S' \to B'$ with the recovery changes applied. Diagram \ref{eq:nt_realiz} expresses a natural transformation as defined in Section \ref{sec:basis}. There are two maps $\delta, \delta'$ between the realization category and the realization category \emph{recovered}. The natural transformation $\alpha$ relates these two maps.

\begin{equation}\label{eq:nt_realiz}
\begin{tikzcd}[column sep=huge, row sep=huge]
Realization
  \arrow[bend left=50]{r}[name=Realization,label=above:$\scriptstyle\mathrm{\delta}$]{}
  \arrow[bend right=50]{r}[name=Realization',label=below:$\scriptstyle \delta'$]{} &
Realization'
  \arrow[shorten <=10pt,shorten >=10pt,Rightarrow,to path={(Realization) -- node[label=right:$\alpha$] {} (Realization')}]{}
\end{tikzcd}
\end{equation}

The implications of the natural transformation are decomposed in Diagram \ref{eq:nt_realiz_comm}. A valid recovery is the one that makes this diagram commute. 

\begin{equation}\label{eq:nt_realiz_comm}
\begin{tikzcd}[row sep=huge]
S(r) \arrow[d, "\delta"'] \arrow[r, "\alpha_{str}", dashed] & S(r') \arrow[d, "\delta'"] \\
B(r) \arrow[r, "\alpha_{beh}", dashed]                   & B(r')                  
\end{tikzcd}
\end{equation}

Marathon 2 recoveries described in Section \ref{sec:recoveries} deal with crowded environments and lack of localization confidence. According to our CT model, this affects the behavioral model. For example, when the planner cannot compute a path because the map is ``full of obstacles'', no component is damaged. However, the solution to this recovery, clearing the global map, may affect the local map model and the controller component. In this case, a change in the behavioral model $\alpha_{beh}$ produces a recovered structure model $S(r')$ through the map $d'^{opp}$ that sends behaviors to structures. 

Another possible contingency is a failure in the laser sensor; in this case, its functionality can be provided by an alternative component, such as the depth camera. In this scenario, the system requires an additional element to transform the point cloud output produced by the depth camera into a laser scan message. Therefore, a contingency in the structural model affects the behavior of the system by changing the information flow. According to Diagram \ref{eq:nt_realiz_comm}, a change in the structural model $\alpha_{str}$ produces a recovered behavioral model $B(r')$ through the map $\delta'$.

\section{Discussion on CT-driven System Modeling}\label{sec:discussion}

System models constitute a vehicle for the realization of systems. They can be used throughout its entire life-cycle; for example, to define requirements, establish a design, analyze it, and verify and validate it. Formalisms can help engineers\textemdash and the robot itself\textemdash identify potential problems, examine complex systems, and communicate ideas with others in a clear and unambiguous way. Moreover, they provide consistency and systematization to increase the reusability of the model. As CT is ``the language'' of abstract mathematics, it becomes the perfect approach to represent systems at different levels of abstraction.

Applied CT has become an increasingly attractive field, as evidenced by the notable growth in publications and research. Although it has a steep learning curve, it can be smoothed out with practitioners and collaborations across domains. The main concern when CT is applied in robotics is the lack of tool support. There is a framework under development, \emph{Catlab}\footnote{\url{https://github.com/AlgebraicJulia/Catlab.jl}}, for applied and computational CT written in the Julia language. However, we believe that CT models can be seen as meta-models that provide a robust formal foundation to OWL ontologies, widely used in robotics~\cite{Olivares-Alarcos2019}. To operationalize them, we shall produce user-friendly interfaces for developers to ease the use of CT models in applied domains.

There are several potential research directions that could be pursued in this area. On the one hand, examine other paradigmatic use-cases in robotics from the CT perspective can be a powerful way to refine model abstractions and show practitioners the value of this approach. The EU-funded ROBOMINERS and CORESENSE projects try to do this. Another promising approach could be the application of the discrete-time process operad formalism~\cite{rupel2013}, to model and analyze system dynamics.

In our case, our next actions will be (i) measure the impact of CT model exploitation in terms of how it affects robot performance and mission fulfillment in real-world scenarios, which we have already addressed from the ontological perspective~\cite{aguado2021}; and (ii) extend the model to other activities in the system life-cycle, such as mission specification, performance metrics, testing, \emph{etc}. These concepts can be used for validation using compositional viewpoint equations, as suggested by~\cite{Schweiker2015, Breiner2020}.

\section{Conclusions}

In this article, we propose the use of formal abstract theories\textemdash Category Theory\textemdash for modeling in robotics. We suggest an approach to help system engineers in the initial phases of understanding CT. Furthermore, we illustrate how these formal tools can provide a systematic approach to addressing the situations that a system engineer may encounter. This is of special importance when building robust and adaptive robots that are systems of extreme complexity.

Following this approach, we subsequently apply the proposed concepts to model the adaptive robot experiment outlined in~\cite{macenski2020}. This system serves as a fundamental yet impactful use case of the \emph{Nav2} stack, widely extended to control mobile robots within the ROS community.

The formal conceptualization goes beyond the engineering phases and endows robots with tools for better understanding of situations, which is a powerful asset towards achieving robustness and resiliency. We propose an approach to leverage explicit engineering knowledge during robot operation, which enhances the robot's dependability. In particular, we model system recoveries as (i) functorial mappings between system realizations, and (ii) natural transformations between structural and behavioral models.

However, there are still many open issues with respect to a full use of CT models in robotic systems. The two main problems that we shall confront are the steep learning curve of these methods, and the lack of engineering-grade tool support. We contemplate the potential use of these formalizations as ontological meta-models, widely used in robotics nowadays. In future work, we aim to extend this model-based approach to other aspects of the robot system life-cycle.

In conclusion, we believe that the steps shown in this article toward formalizing designs and its recoveries are a promising method to enhance robot autonomy. If we have learned something from the past of science and technology, is that effective scientific conceptualizations of real-world phenomena open enormous possibilities for new technology.


\section*{ACKNOWLEDGMENTS}

This work was partially supported by the ROBOMINERS project with funding from the European Union’s Horizon 2020 Research and Innovation Programme (Grant Agreement No. 820971), by the CORESENSE project with funding from the European Union’s Horizon Europe Research and Innovation Programme (Grant Agreement No. 101070254), and by a grant from \textit{Programa Propio} of Universidad Politécnica de Madrid.

\bibliographystyle{abbrv}
{\footnotesize
\bibliography{references}}

\end{document}